\documentclass{INTERSPEECH2023}
\usepackage{stackengine}
\usepackage{CJKutf8}
\usepackage{verbatim}
\usepackage[bottom]{footmisc}

\interspeechcameraready

\title{Incorporating L2 Phonemes Using Articulatory Features \\for Robust Speech Recognition}
\name{Jisung Wang, Haram Lee, Myungwoo Oh}

\address{NAVER Cloud, South Korea}
\email{\{jisung.wang, haram.lee, myungwoo.oh\}@navercorp.com}

\begin{document}

\begin{CJK}{UTF8}{}
\CJKfamily{mj}
\maketitle
 
\begin{abstract}
The limited availability of non-native speech datasets presents a major challenge in automatic speech recognition (ASR) to narrow the performance gap between native and non-native speakers. 
To address this, the focus of this study is on the efficient incorporation of the L2 phonemes, which in this work refer to Korean phonemes, through articulatory feature analysis.
This not only enables accurate modeling of pronunciation variants but also allows for the utilization of both native Korean and English speech datasets.
We employ the lattice-free maximum mutual information (LF-MMI) objective in an end-to-end manner, to train the acoustic model to align and predict one of multiple pronunciation candidates.
Experimental results show that the proposed method improves ASR accuracy for Korean L2 speech by training solely on L1 speech data. 
Furthermore, fine-tuning on L2 speech improves recognition accuracy for both L1 and L2 speech without performance trade-offs.
\end{abstract}
\noindent\textbf{Index Terms}: articulatory features, non-native speech, LF-MMI, robust speech recognition

\section{Introduction}
Automatic speech recognition (ASR) has made tremendous advancements in recent years \cite{Li22-RECENTASR}, but it still struggles to accurately recognize non-native speech, particularly for those whose first language (L1) is significantly different from the target language \cite{Koeneckea20-RACIALDISPARITY}. 
The increasing demand for ASR in non-native speech \cite{Crystal03-ENGLISH} emphasizes the need for more robustness of speech recognition system. 
Research in accent-robust ASR has proposed various methods to improve recognition accuracy, such as model adaptation with transfer learning  \cite{Shibano21-ASRTFL}, pronunciation modeling \cite{Lehr14-PMDIALECT, Goronzy04-LEXADAPT}, and multi-task learning \cite{Prasad20-ACCENTINFO, Huang14-MULTIACCLAYER}. 
However, in-depth studies on improving ASR performance for non-native speakers, especially for languages with scarce speech data resources like Korean, are lacking.

Our research focuses on modeling pronunciation variants specific to Korean English speech using an extended phoneme inventory that incorporates L2 phonemes. 
This is motivated mainly by the significant differences between Korean and English phoneme inventories, and aims to improve recognition accuracy by modeling these differences. 
Note our approach stands out from recent works \cite{Radford22-WHISPER,Aksenova22-ACCASR} by tackling the problem from a linguistic perspective instead of relying on large models or large amounts of data. 
Our study makes several important contributions to ASR for non-native speakers. 
First, our approach enables cost-effective training using only L1 speech data. 
This is achieved through the use of a unified phoneme set that incorporates L2 phonemes, allowing for improved performance without the need for L2 speech data. 
Second, our approach resolves the trade-off issue where the performance of L1 speech decreases as more L2 speech data is added.
Third, this study is the first to investigate and utilize phonological errors specific to Korean English speakers in the context of ASR. 
Finally, we are able to train for multiple pronunciation candidates in an end-to-end manner, eliminating the need for HMM-GMM training and alignment procedures commonly used in pronunciation variants modeling.

The remainder of this paper is organized as follows.
Section~\ref{section:related_work} reviews related work on pronunciation variants modeling.
The proposed approach is explained in detail in Section~\ref{section:methods}, including the preparation of phoneme inventory and pronunciation lexicons and training methods.
In Section~\ref{section:experiments}, we describe the datasets and training procedures, and present the experimental setup.
Experimental results and analysis of the performance of our approach are presented in Section~\ref{section:results}.
Finally, Section~\ref{section:conclusion} concludes our work.

\section{Related work}
\label{section:related_work}
Our research is one of pronunciation variants acoustic modeling techniques with augmented lexicons. 
This approach has been studied in both ASR and mispronunciation detection and diagnosis (MDD) research \cite{Wang22-MANDARINASR, Long21-MANDENGCS, Duan20-CLTL, Yan20-ANTIPHONE, Korzekwa21-MDUNCERTAINTY, Doremalen10-ERRPATTERNMD, Goronzy04-LEXADAPT}.

However, many previous studies have only used the original native phoneme set and have not expanded it to include phonemes specific to non-native speakers, which restricts the model's ability to train on L2 language data.
For example, some studies have augmented the Mandarin lexicon with pronunciation variants, but without incorporating new phonemes relevant to L2 speech \cite{Wang22-MANDARINASR, Long21-MANDENGCS}.
In the meantime, another study has employed L2 phonemes to enable training on larger native speech datasets \cite{Duan20-CLTL}. 
But their approach utilizes separate language-dependent output layers, which limits the use of valuable L2 speech training data when available.
Additionally, a study on expanding the phoneme set for acoustic modeling in the context of MDD has been conducted \cite{Yan20-ANTIPHONE}.
However, the generated phoneme set in that study is simply a duplication of the original inventory with anti-phones and lacks linguistic knowledge, making it unable to train on the speech of L2 speakers' native language \cite{Yan20-ANTIPHONE}.
Moreover, this approach does not handle multiple possible answers during training and randomly assigns anti-phones to L2 speech data labels, which may force the model to learn an incorrect phoneme class. 
Moreover, the approach does not handle multiple possible answers during training and instead randomly assigns anti-phones to L2 speech data labels.
This assignment of anti-phones may force the model to learn a phoneme class, which may not be the correct answer.
In contrast, our approach trains on extended labels and learns a phoneme from multiple hypotheses automatically, without the need for randomly selecting one possible answer.

\section{Methods}
\label{section:methods}
\subsection{Incorporating L2 phonemes} 
\label{subsection:incorporating_l2_phonemes}

\begin{table}[b!]
  \caption{Phonemes of each language which are tied according to their similarity of articulatory features (AFs) are represented in IPA symbols. The symbols in the parentheses of the first column are from CMUDict without stress marks. The second column includes Korean jamo symbols, which are used to represent sounds in the Korean language and are combined to form syllable blocks. \textipa{/k/}, \textipa{/p/}, \textipa{/t/} and \textipa{/\textteshlig/} are typically aspirated in syllable-onset. Note that unlike English vowels, Korean vowels are not distinguished by tense or lax quality.}
  \label{tab:tying_map}
  \centering
  \begin{tabular}{lll}
    \toprule
    \textbf{Eng.}    & \textbf{Kor.}    & \textbf{Common AFs}     \\
    \midrule
    Consonant \\
    \midrule
    k (K)  & k\textsuperscript{h} (ㅋ)                    & velar, plosive, voiceless, \\
                             & &              aspirated (/k/ in syllable-onset) \\
    p (P) &  p\textsuperscript{h} (ㅍ)                   & bilabial, plosive, voiceless, \\ 
                             & &              aspirated (/p/ in syllable-onset) \\
    t (T) & t\textsuperscript{h} (ㅌ)                    & alveolar, plosive, voiceless, \\
                             & &              aspirated (/t/ in syllable-onset) \\
    \textipa{\textteshlig} (CH) & \textipa{t\textctc}\textsuperscript{h} (ㅊ) & postalveolar, affricative, voiceless,   \\
                             & &              aspirated (\textipa{\textteshlig} in syllable-onset) \\
    h (HH)                   & h (ㅎ)                    & glottal, fricative, voiceless, \\
                             & &    aspirated \\
    m (M)                    & m (ㅁ)                    & bilabial, nasal, voiced, neutral\\
    n (N)                    & n (ㄴ)                    & alveolar, nasal, voiced, neutral\\
    \textipa{ŋ} (NG)         & \textipa{ŋ} (ㅇ)         & velar, nasal, voiced, neutral \\
    s (S)                    & s (ㅅ)                     & alveolar, fricative, voiceless, \\
                             & & neutral  \\
    \midrule
    Vowel \\
    \midrule
    \textipa{i} (IY)   & \textipa{i} (ㅣ) & high, front, unrounded  \\
    \textipa{e} (EH)   & \textipa{\textlowering{e}} (ㅔ, ㅐ) & mid, front, unrounded  \\
    \textipa{u} (UW)   & \textipa{u} (ㅜ) & high, back, rounded\\
    \textipa{\textturnv} (AH)  & \textipa{\textsubrhalfring{\textturnv}} (ㅓ) & low, central, unrounded \\
    \bottomrule
  \end{tabular}
\end{table}

The conventional English phonemes, such as those found in CMUDict\footnote{http://www.speech.cs.cmu.edu/cgi-bin/cmudict}, are not comprehensive enough to accurately represent the pronunciations of non-native speakers, particularly Korean speakers. 
This is due to the limited similarity between the phonetic sounds of English and Korean, which makes it challenging for Korean learners to perceive and produce certain sounds.
As a result, Korean speakers often rely on their own phoneme inventory and produce sounds that differ from the actual English sounds.
As an example, consider the word "ring". 
Korean speakers may mispronounce it as \textipa{/l i ŋ/} instead of \textipa{/r \textsci} \textipa{ŋ/}. 
While the English phoneme set includes the sounds \textipa{/l/}, \textipa{/i/}, and \textipa{/ŋ/}, some Korean speakers may also say \textipa{/\textfishhookr} \textipa{i ŋ/} with \textipa{/\textfishhookr/}, which is not part of the English inventory.

Therefore, in order to capture the rich sounds of both L1 and L2 English speech produced by Korean speakers, it is crucial to include additional Korean phonemes. 
However, to avoid redundancy within the phoneme set and reduce the burden on the training system, only phonemes that are not originally part of the English inventory were incorporated. 
To accomplish this, we utilized a process of phoneme tying, where similar phonemes were grouped together.

Firstly, we defined a 36-sized Korean phoneme set, consisting of 19 consonants and 17 vowels, and a 39-sized English phoneme set, consisting of 24 consonants and 15 vowels. 
We then analyzed the articulatory features of each phoneme to determine whether it shared the same or similar features as phonemes in the L1 inventory, and only incorporated L2 phonemes that were distinct from L1 phonemes. 
Our analysis was based on the following categories: place, manner, voicing, aspiration for consonants, and height, frontness, rounding, and tenseness for vowel phonemes.

Table \ref{tab:tying_map} shows the 9 consonants and 4 vowels that were considered to be the same phonemes, resulting in the development of a unified phoneme inventory consisting of 34 consonants and 28 vowels, for a total of 62 phonemes. 
It should be noted that this unified phoneme set allows for training acoustic models on Korean language speech as well, as it includes all the phonemes necessary to represent Korean vocabulary.

\subsection{Phonological errors in language transfer} 
\begin{table}[b!]
  \caption{Major phonological transfers commonly observed in Korean speakers of English. These are not exhaustive and there are other rules not included in this table.}
  \label{tab:negative_transfer}
  \centering
  \begin{tabular}{lll}
    \toprule
    \textbf{L1}    & \textbf{Negative Transfer}    & \textbf{Example}     \\
    \midrule
    Consonant \\
    \midrule
    \textipa{/\dh/} &  \textipa{/t/ /d/} & the: \textipa{/\dh} \textipa{\textschwa/} $\rightarrow$ \textipa{/t} \textipa{\textturnv/} \\
    \textipa{/r/}   &  [del] \textipa{/\textturnv/} \textipa{/l/} \textipa{/\textfishhookr/} & card: \textipa{/k a r d/} $\rightarrow$ \textipa{/k} \textipa{\textturna} \textipa{t/} \\
    \textipa{/\textyogh/} & \textipa{/t\textctc/} \textipa{/}\stackunder[1pt]{t}{\textprimstress\textprimstress}\textipa{\textctc/} & jam: \textipa{/\textyogh} \textipa{\ae} \textipa{m/} $\rightarrow$ \textipa{/\textipa{t\textctc}} \textipa{e m/}\\
    \textipa{/v/}   &  \textipa{/b/ /p/} \textipa{/}\stackunder[1pt]{p}{\textprimstress\textprimstress}\textipa{/} & van: \textipa{/v \ae} \textipa{n/} $\rightarrow$ \textipa{/b e n/} \\
    \midrule
    Vowel \\
    \midrule
    \textipa{/\textopeno/} & \textipa{/o/} \textipa{/\textturnv/} & all: \textipa{/\textopeno} \textipa{l/} $\rightarrow$ \textipa{/o l/} \\
    \textipa{/o\textupsilon/} & \textipa{/o/} & boat: \textipa{/b } \textipa{o\textupsilon} \textipa{t/} $\rightarrow$ \textipa{/p o t/} \\
    \textipa{/\textsci/} & \textipa{/i/} & it: \textipa{/\textsci} \textipa{t/} $\rightarrow$ \textipa{/i t/} \\
    \textipa{/\textupsilon/} & \textipa{/u/} & hood: \textipa{/h} \textipa{\textupsilon} \textipa{d/} $\rightarrow$ \textipa{/h u d/} \\
    \bottomrule
    \end{tabular}
\end{table}

\begin{figure}[t]
  \centering
  \includegraphics[width=\linewidth]{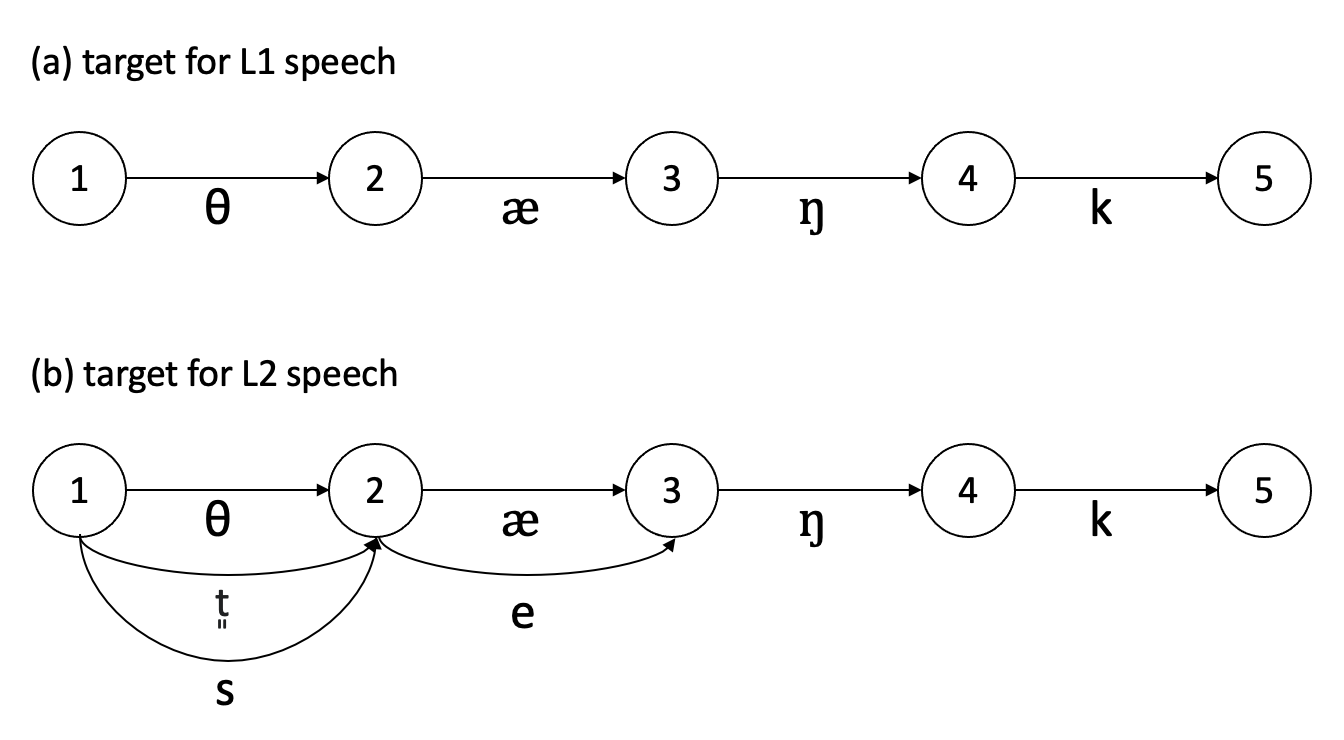}
  \caption{Simplified graphs for the sample word "thank". (a) includes a single phoneme sequence while (b) contains multiple variants of phoneme sequences for "thank" which are generated based on the transfer rules.}
  \label{fig:graph_thank}
\end{figure}

Negative transfer in language refers to the phenomenon where a learner's mother language negatively influences their acquisition of target language. 
One of the most common manifestations is the mispronunciation of the target phonemes that do not exist in the their mother language. 
For instance, Korean-speaking learners often substitute \textipa{/o/} for \textipa{/o\textupsilon/}, \textipa{/v/} to \textipa{/b/}.
Based on foreign language notation \cite{KRMINSTRY17-NOTATION} and previous research on second language acquisition by Korean speakers \cite{Cho06-KORENG}, we developed phonological transfer rules. 
As an illustration, Table~\ref{tab:negative_transfer} shows several transfer rules for consonants and vowels.
With these rules we generated multiple phoneme sequences for each word and expanded our lexicon. 
It is important to note that pronunciation errors can vary widely depending on the speaker's level of proficiency in English and their specific vulnerable points. 
For instance, when pronouncing the English word "thank" \textipa{/\texttheta} \textipa{\ae} \textipa{ŋ} \textipa{k/}, a learner may correctly produce the phoneme \textipa{/\texttheta/} but struggle with \textipa{/\ae/}. 
To model the various combinations of pronunciations that may occur, we used the open-source toolkit OpenFST\footnote{https://www.openfst.org} to encode them in the format of finite-state transducers (FSTs). 
Figure~\ref{fig:graph_thank} (b) illustrates a simple graph that encodes six possible phoneme sequences for the word "thank", with three phoneme options for \textipa{/\texttheta/} and two for \textipa{/\ae/}.

\subsection{End-to-end LF-MMI training for multiple candidates} 
\label{subsection:training}
Training acoustic model on multiple candidates as targets (Figure~\ref{fig:graph_thank} (b)) is challenging even with advanced speech learning techniques such as connectionist temporal classification \cite{Graves06-CTC} or recurrent neural network transducer \cite{Graves12-RNNT}.
To address this, we use the LF-MMI objective with multi-answer lexicon FSTs.
The LF-MMI objective function \cite{Povey16-LFMMI} for given speech \(\mathbf{X}^{(u)}\) is defined as:
\begin{align}
\mathcal{F} = log\frac{p(\mathbf{X}^{(u)}|\mathbb{G}_{num^{(u)}})}{p(\mathbf{X}^{(u)}|\mathbb{G}_{den})}
\end{align}
where \(\mathbb{G}_{num^{(u)}}\) is the numerator graph which includes phoneme paths for target utterance $u$, and \(\mathbb{G}_{den}\) is the denominator graph that represents all possible phoneme sequences.
As the LF-MMI objective function maximizes the probability of the target sequence for the given speech while minimizing the probability of other labels, incorporating multiple answers within the numerator graph using augmented lexicon FSTs enables training for multiple candidates.

In practice, we use a composite HMM with self-loops as the numerator graph. 
The self-loops without any restrictions allow the model to learn alignments freely, enabling end-to-end training. 
To set the target labels for our acoustic model, we use a 1-state HMM topology and a tree-free context-dependent biphone. 
These settings transform the 63 phoneme classes from our unified inventory (Section~\ref{subsection:incorporating_l2_phonemes}), which includes additional silence phoneme, into a set of 4,032 (= 63 $\times$ 64) context-dependent states.

\section{Experiments}
\label{section:experiments}
\subsection{Datasets}
\label{subsection:datasets}

\begin{table}[b!]
  \caption{Dataset description.}
  \label{tab:dataset}
  \centering
  \begin{tabular}{llll}
    \toprule
    \textbf{L1 / L2} & \textbf{Name}    & \textbf{hours}    & \textbf{\# utt}     \\
    \midrule
    \textbf{Train} \\
    \midrule
    L1  & LibriSpeech & 960 & 281,241 \\
        & KsponSpeech (Korean) & 960 & 616,630 \\
    L2  & EngDictKr 100h & 100 & 75,495 \\
        & EngDictKr 1000h & 1,000 & 774,994 \\
    \midrule
    \midrule
    \textbf{Test} \\
    \midrule
    L1  & LS dev-clean/other  & 5.39/5.12 & 2,703/2,864 \\
        & LS test-clean/other & 5.34/5.40 & 2,620/2,939 \\
        & CMU-ARCTIC          & 4 & 4,524 \\
    L2  & L2-ARCTIC-Kr        & 4 & 4,524 \\
    \bottomrule
    \end{tabular}
\end{table}

We employed three types of datasets: English speech uttered by native speakers, English speech spoken by Korean speakers, and a dataset of Korean speech. Specifically, we utilized LibriSpeech (LS) \cite{Panayotov15-LIBRISPEECH} and CMU-ARCTIC \cite{Kominek03-L1ARCTIC} as the English corpus spoken by natives, L2-ARCTIC \cite{Zhao18-L2ARCTIC} and our in-house EngDictKr as the dataset of English speech spoken by Koreans. Finally, KsponSpeech \cite{Bang20-KSPONSPEECH} was used as the dataset of Korean speech. (See details in Table~\ref{tab:dataset}.) 

The L2-ARCTIC dataset, which includes L2 English speech from 24 non-native speakers who speak 6 different L1 languages, was used for evaluation purposes only.
We exclusively utilized speech data produced by four Korean speakers, as our study's primary goal is to enhance the recognition accuracy for Korean speakers.
For additional evaluation, we used CMU-ARCTIC, the L1 data that L2-ARCTIC is based on, including speech from four speakers.
To evaluate the effectiveness of utilizing L2 speech training data, we created a dataset called EngDictKr through crowd-sourcing.
This dataset contains 1,000 hours of English speech read by Korean speakers using mobile devices.
The sentences used in the dataset were extracted from several dictionaries, including Oxford English-Korean dictionary \cite{Jung09-OXFORDDICT}.
The users who participated in the data collection had varying degrees of proficiency in English pronunciation.
We used this non-public training data due to the lack of publicly available Korean English speech sets. 

\subsection{Experimental setup}
\label{subsection:experimental_setup}

\begin{table*}[t]
\caption{Word Error Rate (WER) (\%) of evaluation sets tested on models trained from scratch.} 
\label{tab:wer_base}
\centering
\setlength{\tabcolsep}{5pt}
\begin{tabular}{c|c|cccccc}
\toprule
\textbf{\#} & \textbf{train set}       & \textbf{dev-clean}  &  \textbf{dev-other}  &  \textbf{test-clean} & \textbf{test-other} & \textbf{CMU-ARCTIC} & \textbf{L2-ARCTIC-Kr} \\
\midrule
1 & LS &  \textbf{3.77}    &  8.93  &  \textbf{4.21}  &  \textbf{9.27} & \textbf{2.59} & 10.75  \\
2 & LS + EngDictKr 100h & 3.83 & \textbf{8.85} & 4.27 & 9.33 & 2.80 & 9.76 \\
3 & LS + KsponSpeech (proposed) &  3.95  &  8.98  &  4.45  & 9.50 & 2.67 & \textbf{9.67} \\
\bottomrule
\end{tabular}
\end{table*}

\begin{table*}[t]
\caption{WER (\%) of evaluation sets tested on fine-tuned models with EngDictKr data, from pre-trained models. Models \#1 and \#3 were fine-tuned from model \#1 in Table~\ref{tab:wer_base}. Models \#2 and \#4 were fine-tuned from model \#3 in Table~\ref{tab:wer_base}.}
\label{tab:wer_ft}
\centering
\setlength{\tabcolsep}{5pt}
\begin{tabular}{c|cc|cccccc}
\toprule
\textbf{\#} & \textbf{num. of phonemes} & \textbf{L2 hours} & \textbf{dev-clean} & \textbf{dev-other} & \textbf{test-clean} & \textbf{test-other} & \textbf{CMU-ARCTIC} & \textbf{L2-ARCTIC-Kr} \\
\midrule
1 & 39 & 100 & 3.78 & 8.61 &  4.27  &  9.00 & 2.58 & 9.59   \\
2 & 62 (proposed) & 100 & 3.80  &  8.20   & 4.23  & 8.86  & 2.66 & 8.72 \\
\midrule
3 & 39 & 1,000 & 3.94 & 8.95 & 4.44 & 9.47 & 2.81 & 9.56 \\
4 & 62 (proposed) & 1,000 & \textbf{3.74} & \textbf{7.84} & \textbf{4.11} & \textbf{8.36} & \textbf{2.48} & \textbf{8.59} \\
\bottomrule
\end{tabular}
\end{table*}

Our model is a conformer encoder architecture \cite{Gulati20-CFM} without a decoder network. 
The model is 26.8M parameter sized, comprising of 16 layers with 256 dimensions and 8 multi-head attentions. 
80-dimensional filterbank features are computed from a 25ms window with a step size of 10ms and fed into the convolution subsampling layers to achieve 40ms rate.
The encoder network is trained on full-biphones, resulting in 4,032 classes (as explained in Section~\ref{subsection:training}).
For decoding, we used a WFST graph composed of a tri-gram language model, which was subsequently rescored using a 4-gram model.
The LM training and decoding processes were done by following Kaldi \cite{Povey11-KALDI} recipe for LS.
We trained the model using the Adamw optimizer \cite{Loshchilov19-ADAMW} with parameter values of $\beta_{1}=0.9$, $\beta_{2}=0.999$, and $\varepsilon=10^{-9}$. 
After 10 epochs, the learning rate is reduced from the initial value of 2e-4 when the validation set's loss did not improve with a patience of 1. 
We chose 184 batch size and LS dev-clean set as validation set throughout experiments. 
Finally, our training objective is an end-to-end version of LF-MMI \cite{Hadian18-E2ELFMMI}, which removes the dependency on HMM-GMM modeling and enables us to train on speech with multiple possible pronunciations in a single stage.
LF-MMI training was done using a tool pychain \cite{Shao20-PYCHAIN}, which enables full GPU training on both numerator and denominator graphs (Section~\ref{subsection:training}).

\section{Results}
\label{section:results}
Table~\ref{tab:wer_base} displays the word error rate (WER) for our evaluation sets, which consist of both L1 (LibriSpeech dev-clean/other, test-clean/other, CMU-ARCTIC) and L2 (L2-ARCTIC-Kr) speech datasets.
We found that the model \#3, trained on both LibriSpeech and KsponSpeech, resulted in a significant 10.0\% relative improvement in WER for the L2-ARCTIC-Kr, compared to the baseline model (\#1) that was trained solely on LibriSpeech. 
It is worth noting that model \#3 was not exposed to any L2 speech during training, yet it still achieved an improvement in recognition for L2 speech. 
Furthermore, the WER of 9.67\% for L2 speech was similar to or lower than the WER of 9.76\% obtained by model \#2, which was trained on EngDictKr 100h and LibriSpeech without our proposed method. 
This outcome can be attributed to the fact that the model was trained on Korean phonemes that are present in the KsponSpeech data. 
Although recognition rate for L1 speech sets are slightly worse than the models \#1 and \#2, these results demonstrate that the Korean phonemes from KsponSpeech were successfully incorporated into the model, highlighting the effectiveness of our approach for improving speech recognition for non-native speakers even in the absence of L2 speech. 
This is an important finding since obtaining L2 speech data is typically challenging, while L1 speech datasets are relatively easy to obtain. 

Table~\ref{tab:wer_ft} shows the WER results for the fine-tuned models trained with EngDictKr and LibriSpeech data.
Models \#1 and \#2 were fine-tuned on 100 hours of EngDictKr, while models \#3 and \#4 were fine-tuned on 1,000 hours of EngDictKr.
Fine-tuning on 100 hours of EngDictKr resulted in improvements for L2 speech recognition, with or without our proposed method (models \#1 and \#2).
However, the model trained with our proposed method (\#2) achieved better performance for L2 speech (8.72\%) compared to the model without our proposed method (\#1) which had a WER of 9.59\%.
Furthermore, after fine-tuning a model with a substantial amount of 1,000 hours of EngDictKr data using our proposed method (\#4), it achieved the lowest WER of 8.59\% for L2 speech and also attained the lowest WERs for all the L1 speech evaluation sets.
In contrast, model \#3, fine-tuned with the same 1,000 hours of EngDictKr data but without our proposed method, resulted in a higher WER of 9.56\% for L2 speech. 
Model \#2, fine-tuned with only 100 hours of EngDictKr but with our proposed method, outperformed it with a WER of 8.72\% for L2 speech.
In fact, model \#3 showed performance trade-offs between L1 and L2 speech, with WERs for L1 speech increasing while a slight improvement in WER for L2 speech was observed.
This degradation in L1 speech recognition is believed to be due to training on incorrect pronunciation, which may lead to the grouping of different phonemes incorrectly.
For instance, training on \textipa{/θ/} and \textipa{/s/} for a single class \textipa{/θ/} could harm the model's L1 speech recognition performance if a large number of substitutions of \textipa{/s/} pronunciation occurs in the training dataset.
In contrast, model \#4, fine-tuned with our proposed method, showed no such trade-offs and achieved the lowest WERs for both L1 and L2 speech evaluation sets.

Additionally, we evaluated L2 speech from speakers of non-Korean languages (the rest of L2-ARCTIC).
As Table~\ref{tab:wer_others} shows, the WER has decreased for these speakers, despite the absence of language-specific modeling for them.
This implies potential for further improvements if variations specific to these languages are modeled.
And it also indicates robustness against potential imprecise modeling of variations.
\begin{table}[t]
\caption{WER (\%) of CMU-ARCTIC (L1) and L2-ARCTIC-Others (L2), excluding Korean speakers from L2-ARCTIC.}
\label{tab:wer_others}
\centering
\setlength{\tabcolsep}{5pt}
\begin{tabular}{c|cc}
\toprule
\textbf{train set}       & \textbf{L1}  &  \textbf{L2} \\
\midrule
LS &  2.59 & 16.25  \\
LS + EngDictKr & 2.81 & 17.16 \\
LS + EngDictKr (proposed) &  \textbf{2.48} & \textbf{15.60} \\
\bottomrule
\end{tabular}
\end{table}

\section{Conclusion}
\label{section:conclusion}
In this paper, we proposed an approach to improve speech recognition accuracy for non-native speakers by modeling pronunciation variants specific to Korean English speech using an extended phoneme inventory. 
We incorporated L2 phonemes based on articulatory feature analysis and employed an end-to-end training approach for multiple answers, created based on pre-defined transfer rules.
Our proposed methods meaningfully increased the recognition accuracy of Korean English with only L1 speech. 
Furthermore, fine-tuning on a comparable amount of L2 speech led to significant improvements for both L1 and L2 speech while the other experiments without our methods showed performance trade-offs between L1 and L2 speech.
Lastly, the observed improvements in L2 speech from speakers of non-target languages suggest that our approach holds potential for broader applicability across other languages.
For future work, we could explore using self-supervised pre-trained models as a strong baseline to enhance acoustic feature learning, potentially improving our approach's performance.

\bibliographystyle{IEEEtran}
\bibliography{mybib}

\begin{thebibliography}{10}
\providecommand{\url}[1]{#1}
\csname url@samestyle\endcsname
\providecommand{\newblock}{\relax}
\providecommand{\bibinfo}[2]{#2}
\providecommand{\BIBentrySTDinterwordspacing}{\spaceskip=0pt\relax}
\providecommand{\BIBentryALTinterwordstretchfactor}{4}
\providecommand{\BIBentryALTinterwordspacing}{\spaceskip=\fontdimen2\font plus
\BIBentryALTinterwordstretchfactor\fontdimen3\font minus
  \fontdimen4\font\relax}
\providecommand{\BIBforeignlanguage}[2]{{%
\expandafter\ifx\csname l@#1\endcsname\relax
\typeout{** WARNING: IEEEtran.bst: No hyphenation pattern has been}%
\typeout{** loaded for the language `#1'. Using the pattern for}%
\typeout{** the default language instead.}%
\else
\language=\csname l@#1\endcsname
\fi
#2}}
\providecommand{\BIBdecl}{\relax}
\BIBdecl

\bibitem{Li22-RECENTASR}
J.~Li \emph{et~al.}, ``Recent advances in end-to-end automatic speech
  recognition,'' \emph{APSIPA Transactions on Signal and Information
  Processing}, vol.~11, no.~1, 2022.

\bibitem{Koeneckea20-RACIALDISPARITY}
A.~Koeneckea, A.~Namb, E.~Lakec, J.~Nudelld, M.~Quarteye, Z.~Mengeshac,
  C.~Toupsc, J.~R. Rickfordc, D.~Jurafskyc, and S.~Goel, ``Racial disparities
  in automated speech recognition,'' in \emph{Proc. the National Academy of
  Sciences}, vol. 117, no.~14, 2020, pp. 7684--7689.

\bibitem{Crystal03-ENGLISH}
D.~Crystal, ``English as a global language,'' \emph{Cambridge university
  press}, 2003.

\bibitem{Shibano21-ASRTFL}
T.~Shibano, X.~Zhang, M.~T. Li, H.~Cho, P.~Sullivan, and M.~Abdul-Mageed,
  ``Speech technology for everyone: Automatic speech recognition for non-native
  {E}nglish,'' \emph{in Proc. the 4th International Conference on Natural
  Language and Speech Processing}, 2021.

\bibitem{Lehr14-PMDIALECT}
M.~Lehr, K.~Gorman, and I.~Shafran, ``Discriminative pronunciation modeling for
  dialectal speech recognition,'' \emph{in Proc. INTERSPEECH}, 2014.

\bibitem{Goronzy04-LEXADAPT}
S.~Goronzy, R.~Kompe, and S.~Rapp, ``Generating non-native pronunciation
  variants for lexicon adaptation,'' \emph{Speech Communication}, vol.~42,
  no.~1, 2004.

\bibitem{Prasad20-ACCENTINFO}
A.~Prasad and P.~Jyothi, ``How accents confound: Probing for accent information
  in end-to-end speech recognition systems,'' \emph{in Proc. the 58th Annual
  Meeting of the Association for Computational Linguistics}, pp. 3739--3753,
  2020.

\bibitem{Huang14-MULTIACCLAYER}
Y.~Huang, D.~Yu, C.~Liu, , and Y.~Gong, ``Multi-accent deep neural network
  acoustic model with accent-specific top layer using the kld-regularized model
  adaptation,'' \emph{Fifteenth Annual Conference of the International Speech
  Communication Association}, 2014.

\bibitem{Radford22-WHISPER}
A.~Radford, J.~W. Kim, T.~Xu, G.~Brockman, C.~McLeavey, and I.~Sutskever,
  ``Robust speech recognition via large-scale weak supervision,'' \emph{arXiv
  preprint arXiv:2212.04356}, 2022.

\bibitem{Aksenova22-ACCASR}
A.~Aks{\"e}nova, Z.~Chen, C.-C. Chiu, D.~van Esch, P.~Golik, W.~Han, L.~King,
  B.~Ramabhadran, A.~Rosenberg, S.~Schwartz, and G.~Wang, ``Accented speech
  recognition: Benchmarking, pre-training, and diverse data,'' \emph{arXiv
  preprint arXiv:2205.08014}, 2022.

\bibitem{Wang22-MANDARINASR}
L.~Wang and R.~Tong, ``{Pronunciation modeling of foreign words for Mandarin
  ASR by considering the effect of language transfer},'' \emph{in Proc.
  INTERSPEECH}, pp. 1443--1447, 2014.

\bibitem{Long21-MANDENGCS}
Y.~Long, S.~Wei, J.~Lian, and Y.~Li, ``Pronunciation augmentation for
  mandarin-english code-switching speech recognition,'' \emph{EURASIP Journal
  on Audio, Speech, and Music}, 2021.

\bibitem{Duan20-CLTL}
R.~Duan, T.~Kawahara, M.~Dantsuji, and H.~Nanjo, ``Cross-lingual transfer
  learning of non-native acoustic modeling for pronunciation error detection
  and diagnosis,'' \emph{IEEE/ACM Transactions on Audio, Speech, and Language
  Processing}, vol.~28, 2020.

\bibitem{Yan20-ANTIPHONE}
B.-C. Yan, M.-C. Wu, H.-T. Hung, and B.~Chen, ``An end-to-end mispronunciation
  detection system for l2 english speech leveraging novel anti-phone
  modeling,'' in \emph{Proc. INTERSPEECH}, 2020.

\bibitem{Korzekwa21-MDUNCERTAINTY}
D.~Korzekwa, J.~Lorenzo-Trueba, S.~Zaporowski, S.~Calamaro, T.~Drugman, and
  B.~Kostek, ``Mispronunciation detection in non-native (l2) english with
  uncertainty modeling,'' in \emph{Proc. IEEE ICASSP}, 2021, pp. 7738--7742.

\bibitem{Doremalen10-ERRPATTERNMD}
J.~van Doremalen, C.~Cucchiarini, and H.~Strik, ``Using non-native error
  patterns to improve pronunciation verification,'' in \emph{Proc.
  INTERSPEECH}, 2010.

\bibitem{KRMINSTRY17-NOTATION}
{Korean Ministry of Culture, Sports and Tourism}, ``Foreign language
  notation,'' \emph{Korean Ministry of Culture, Sports and Tourism Notice}, no.
  2017-14, 2017.

\bibitem{Cho06-KORENG}
J.~Cho and H.-K. Park, ``A comparative analysis of korean-english phonological
  structures and processes for pronunciation pedagogy in interpretation
  training,'' \emph{Meta}, vol.~51, no.~2, pp. 229--246, 2006.

\bibitem{Graves06-CTC}
A.~Graves, S.~Fern{\'a}ndez, F.~Gomez, and J.~Schmidhuber, ``Connectionist
  temporal classification: labelling unsegmented sequence data with recurrent
  neural networks,'' in \emph{Proceedings of the 23rd international conference
  on machine learning}, 2006, pp. 369--376.

\bibitem{Graves12-RNNT}
A.~Graves, ``Sequence transduction with recurrent neural networks,'' in
  \emph{Proceedings of the 29th international conference on machine learning},
  2012.

\bibitem{Povey16-LFMMI}
D.~Povey, V.~Peddinti, D.~Galvez, P.~Ghahremani, V.~Manohar, X.~Na, Y.~Wang,
  and S.~Khudanpur, ``Purely sequence-trained neural networks for asr based on
  lattice-free mmi.'' in \emph{Proc. INTERSPEECH}, 2016, pp. 2751--2755.

\bibitem{Panayotov15-LIBRISPEECH}
V.~Panayotov, G.~Chen, D.~Povey, and S.~Khudanpur, ``Librispeech: An asr corpus
  based on public domain audio books,'' in \emph{Proc. IEEE ICASSP}, 2015, pp.
  5206--5210.

\bibitem{Kominek03-L1ARCTIC}
J.~Kominek and A.~W. Black, ``Cmu arctic databases for speech synthesis,'' in
  \emph{5th ISCA Workshop on Speech Synthesis}, 2003.

\bibitem{Zhao18-L2ARCTIC}
G.~Zhao, S.~Sonsaat, A.~Silpachai, I.~Lucic, E.~Chukharev-Hudilainen, J.~Levis,
  and R.~Gutierrez-Osuna, ``L2-arctic: A non-native english speech corpus,'' in
  \emph{Proc. INTERSPEECH}, 2018, pp. 2783--2787.

\bibitem{Bang20-KSPONSPEECH}
J.-U. Bang, S.~Yun, S.-H. Kim, M.-Y. Choi, M.-K. Lee, Y.-J. Kim, D.-H. Kim,
  J.~Park, Y.-J. Lee, and S.-H. Kim, ``Ksponspeech: Korean spontaneous speech
  corpus for automatic speech recognition,'' \emph{Applied Sciences}, vol.~10,
  2020.

\bibitem{Jung09-OXFORDDICT}
Y.~Jung and M.~Cho, \emph{Oxford Advanced Learner's English-Korean
  Dictionary}.\hskip 1em plus 0.5em minus 0.4em\relax Oxford University Press,
  2009.

\bibitem{Gulati20-CFM}
A.~Gulati, J.~Qin, C.-C. Chiu, N.~Parmar, Y.~Zhang, J.~Yu, W.~Han, S.~Wang,
  Z.~Zhang, Y.~Wu, and R.~Pang, ``Conformer: Convolution-augmented transformer
  for speech recognition,'' \emph{in Proc. INTERSPEECH}, 2020.

\bibitem{Povey11-KALDI}
D.~Povey, A.~Ghoshal, G.~Boulianne, L.~Burget, O.~Glembek, N.~Goel,
  M.~Hannemann, P.~Motlicek, Y.~Qian, P.~Schwarz \emph{et~al.}, ``The kaldi
  speech recognition toolkit,'' in \emph{IEEE 2011 workshop on automatic speech
  recognition and understanding}, no. CONF.\hskip 1em plus 0.5em minus
  0.4em\relax IEEE Signal Processing Society, 2011.

\bibitem{Loshchilov19-ADAMW}
I.~Loshchilov and F.~Hutter, ``Decoupled weight decay regularization,'' in
  \emph{International Conference on Learning Representations}, 2019.

\bibitem{Hadian18-E2ELFMMI}
H.~Hadian, H.~Sameti, D.~Povey, and S.~Khudanpur, ``End-to-end speech
  recognition using lattice-free {MMI},'' in \emph{Proc. INTERSPEECH}, 2018.

\bibitem{Shao20-PYCHAIN}
Y.~Shao, Y.~Wang, D.~Povey, and S.~Khudanpur, ``{PYCHAIN}: A fully parallelized
  pytorch implementation of {LF-MMI} for end-to-end {ASR},'' \emph{in Proc.
  INTERSPEECH}, 2020.

\end{thebibliography}

\end{CJK}
\end{document}